\documentclass[runningheads]{llncs}
\usepackage{graphicx}
\usepackage{amsmath}
\usepackage[algo2e]{algorithm2e} 
\usepackage{algorithmic}
\usepackage{algorithm}
\usepackage[scriptsize]{subfigure}
\usepackage[small]{caption}
\usepackage{array}
\usepackage{caption}
\usepackage{etoolbox}

\AtBeginEnvironment{tabular}{\scriptsize}

\renewcommand{\arraystretch}{1}
\usepackage{rotfloat}
\begin{document}
\title{Adversarial False Data Injection Attack against Nonlinear AC State Estimation with ANN in Smart Grid}
%
%
\author{Tian Liu\and Tao Shu} 
\authorrunning{T. Liu et al.}
\titlerunning{Adversarial FDI Attack against AC State Estimation with ANN}
%
\institute{Department of Computer Science and Software Engineering, Auburn University 
\email{\{tzl0031, tshu\}@auburn.edu}}
\maketitle              
\begin{abstract}
Artificial neural network (ANN) provides superior accuracy for nonlinear alternating current (AC) state estimation (SE) in smart grid over traditional methods. However, research has discovered that ANN could be easily fooled by adversarial examples. In this paper, we initiate a new study of adversarial false data injection (FDI) attack against AC SE with ANN: by injecting a deliberate attack vector into measurements, the attacker can degrade the accuracy of ANN SE while remaining undetected. We propose a population-based algorithm and a gradient-based algorithm to generate attack vectors. The performance of these algorithms are evaluated through simulations on IEEE 9-bus, 14-bus and 30-bus systems under various attack scenarios. Simulation results show that DE is more effective than SLSQP on all simulation cases. The attack examples generated by DE algorithm successfully degrade the ANN SE accuracy with high probability.
\keywords{smart grid \and AC state estimation\and false data injection attack \and adversarial learning}
\end{abstract}
\section{Introduction}
\vspace{-0.2cm}
With the increase of residential and industrial power demand, nowadays a regional or nationwide power outage often leads to catastrophes in the matter of public safety. After the Northeast Blackout of the US in 2003, the US and Canada have reached a consensus in transferring into smart grid system, which is cleaner, more efficient, reliable, resilient and responsive than traditional grid. While the  transition provides many attractive new features such as remote and automatic grid monitoring, control, and pricing, it also raised serious security challenges by opening up traditional power system to many potential attacks in the cyber space. In the 2015 Ukraine power outage (\cite{Liang2017}, \cite{case2016analysis}), the hacker successfully compromised the information systems of three energy distribution companies and caused power disruption to over 225,000 customers. Since then, cyber attacks in smart grid have become a realistic and growing concern. Therefore, it is significant and urgent to identify possible threats, and propose countermeasures to eliminate such threats, so as to reduce the potential loss of the society.

State estimation (SE) plays an important role in system monitoring and control, as it provides current system status for control center operators to take actions in advance to avoid potential accidents. In \textit{alternating current} (AC) power flow systems, states and measurements are non-linearly related, several efforts have been made to adopt artificial neural networks (ANNs) for AC SE to better model this non-linear relationship~(\cite{abdel2018novel}, \cite{mosbah2015multilayer}, \cite{onwuachumba2014new}, \cite{Srivastava}). Although the training phase of these ANN models is costly, it has been shown that these ANN SE models are more accurate. However, studies have shown that SE is vulnerable to \textit{false data injection} (FDI) attacks~\cite{liu2011false}. The adversary can corrupt state variables by injecting a well-coordinated false data to meter measurements, while evading detection. FDI attacks on \textit{direct current} (DC) power flow model can be linearly formulated, hence it is easy to understand their impact and propose countermeasures against them. Nonetheless, FDI attacks towards AC SE are more complicated. The non-linearity between state variables and measurements diffuses the strength of false injection, and makes it hard to identify how the change of measurements would result in the errors to state variables. There are few studies tackling the FDI attacks to AC SE. Jia et al. \cite{jia2012nonlinearity} claims that the ability of malicious attack designed for DC power flow model is alleviated on AC power flow model. Several FDI attacks against AC SE are derived from DC FDI attack and predominantly based upon \textit{weighted least square} (WLS) (\cite{hug2012vulnerability}, \cite{rahman2013false}). 

In the area of image classification, Szegedy first noticed that neural networks may be easily fooled by well-coordinated samples with small perturbations~\cite{szegedy2013intriguing}. Since then, there have been many efforts in exploring the robustness of ANN by designing adversarial attacks (\cite{goodfellow2014explaining}, \cite{Su2017}).

In this paper, we are interested in examining whether the above vulnerability of ANN presented in image classification problem can be extended to SE problem in smart grid. Furthermore, we attempt to develop algorithms that can systematically generate polluted measurements that maximize ANN SE error while eluding from being detected by the bad data detector. By answering these questions, we intend to establish new understanding on the security vulnerabilities of the latest high-accuracy ANN AC SE models. To the best of our knowledge, our work is the first in the literature that studies FDI to ANN AC SE models.

Solving our problem faces new and significant challenges. First, our problem has an optimization nature in the sense that we seek the optimal attack vector that maximizes the attack outcomes, while the goal of the image-classification counterparts is just to find a feasible attack vector. Second, the attack model in our problem considers the attacker's access and resource constraints, by which the attacker only has access to and can only manipulate a certain numbers of meters. In contrast, the image-classification problem has no such constraint and the attacker is allowed to change any pixel of the image. Lastly, the output of ANN SE model is a continuous value, whereas that of the image-classification is discrete and covers a limited number of pre-defined cases. Due to these fundamental structural differences, the existing results from image-classification ANN are not directly applicable to our problem. The main contributions of our work include the following four-fold:
\begin{itemize}
\item In creating the target ANN SE models for large-scale grid systems, a novel penalty term is proposed for the loss function, which significantly improves the accuracy of the ANN on modeling voltage phase angle.
\item An optimization-based FDI attack formulation is proposed for AC SE under the nonlinear ANN model, which can accommodate various practical constraints on the attacker, including its resource and meter accessibility. 
\item We adapt DE and SLSQP to solve the above optimization, targeting at two different attack scenarios: DE generates attack vectors for which the attacker can compromise any $k$ meters, while DE and SLSQP target on the scenario where the attacker is restricted to compromise specific $k$ meters.
\item The effectiveness of the proposed attack models is verified based on extensive simulations on three test systems under various attack scenarios. Our results show that the DE attack succeeds for more than $80\%$ cases even with a small number of compromised meters and low false injection level.  
\end{itemize}

This paper is structured as the following. We start by providing preliminary for SE and bad data detection in Section 2.  We construct AC SE models with ANN as targets for our attacks in Section 3. Subsequently, we introduce our adversary model in Section 4. Our two attack algorithms, DE algorithm and SLSQP algorithm would be presented in Section 5. Finally, the experimental analysis and the comparison of our two attacks on target models and conclusions are presented in Section 6 and Section 7. 
\vspace{-0.3cm}
\section{Preliminaries}
\vspace{-0.1cm}
In AC power flow model, measurements of power flows are non-linearly dependent on state variables, as characterized by the following equations:
$\mathbf{z} = h(\mathbf{x}) + \mathbf{e}$, where $\mathbf{z}$ and $\mathbf{x}$ denote a $N_m$-dimension measurement vector and a $N_n$-dimension state vector, respectively, $\textbf{e}$ denotes a $N_m$-dimension vector of normally distributed measurement errors, and $h$ is a set of nonlinear functions relating states to measurements. In an over-determined case $(N_m > N_n)$, the state variables are determined from WLS optimization over a residual function $J(\mathbf{x})$ \cite{wood2013power}:
\begin{equation}
\hat{\mathbf{x}}=\mathop{\arg\min}_{x}J(\mathbf{x}), \mbox{where } J(\mathbf{x})=(\mathbf{z} - h(\mathbf{x}))^T\mathbf{W}(\mathbf{z} - h(\mathbf{x})) \label{EQ1}
\end{equation}

Here, the weight matrix $\textbf{W}$ is defined as $diag\{\sigma_1^{-2}, \sigma_2^{-2}, ..., \sigma_{N_m}^{-2}\}$, and $\sigma_i^2$ is the variance of the $ith$ measurement ($i=1, ..., N_m$).

Bad measurements would be introduced due to various reasons, such as measuring noise, transmission error, meter malfunction and malicious attack. The ability to detect and identify bad data is extremely critical to the stability of a smart grid. Most bad data detection schemes rely on the residuals $J(\hat{x})$ as their decision variable. In particular, given the assumption that $\mathbf{e}$ is normally distributed, it is shown that $J(x)$ follows $\chi^2(K)$ distribution, where $K=N_m - N_n$ is the degree of freedom \cite{wood2013power}. Any residual exceeding some pre-determined threshold $\tau$ is recognized as a bad data:
\vspace{-0.1cm}
\begin{equation}
\text{$z$ is identified as a bad data, if } J(\mathbf{\hat{x}})=(\mathbf{z} - h(\mathbf{\hat{x}}))^T\mathbf{W}(\mathbf{z} - h(\mathbf{\hat{x}})) > \tau. \label{EQ2} 
\end{equation}

The threshold $\tau$ can be determined by a significant level $\alpha$ in hypothesis testing, by which the false alarms would occur with probability $\alpha$. 
\vspace{-0.3cm}
\section{ANN-based AC SE}
\vspace{-0.2cm}
In lack of actual state-measurement data from real power grid, the training and testing cases in our study are generated based on simulations over the IEEE test systems (9-bus, 14-bus, 30-bus). A Matlab package, MATPOWER \cite{zimmerman2011matpower}, is used for data generation and power flow analysis. Note that the use of simulated data in training does not affect the validity of the proposed ANN model. Usually the model is trained off-line, and then to be retrain or improved with the accumulation of actual data following the same procedure. 

State variables, including magnitudes and phase angles of bus voltages may change within a small range under different loads. To account for this dynamic behavior, we consider a series of loads of the power grid ranging from 80\% to 120\% to simulate actual load pattern. For each instance of the load, the states are calculated by power flow analysis using MATPOWER. According to~\cite{ANSI2008}, $\pm 2\%$ error is allowed in a power measurement reading. In line with this specification, we add an independent Gaussian noise $\epsilon$ to each measurement reading $\psi$, so that the simulated measurement reading becomes $(1+\epsilon)\psi$, where $\epsilon \sim N(0, 0.67\%^2)$. For each of systems, 10,000 and 1,000 state-measurement pairs are generated for training and testing, respectively.

Three ANN SE models are trained for the three systems, respectively. Following \cite{jain2008topological} \cite{abdel2018novel} \cite{menke2018distribution} and \cite{mosbah2015multilayer}, each ANN SE model possesses a multi-layered perceptron architecture. We use mean WLS error as loss function:
\vspace{-0.1cm}
\begin{equation}
loss(\mathbf{z}, \mathbf{x})= \frac{1}{N}\sum_{i=1}^{N}(\mathbf{z} - h(\mathbf{x}))^T\mathbf{W}(\mathbf{z} - h(\mathbf{x})) \label{EQ3}
\end{equation}
\vspace{-0.1cm}
where $N$ is number of training samples. Our experiments show this loss function works well for small-scale systems (such as 9-bus and 14-bus power grid), but fails to provide accurate estimation for voltage phase angles in larger scale systems (30-bus power grid). This is consistent with previous findings in \cite{menke2018distribution}. To address this issue, we revise the loss function in Eq.(\ref{EQ3}) by adding a new penalty term of the \textit{mean square error} (MSE) between the actual and the estimated states, leading to the new loss function in Eq.(\ref{EQ4}) specially designed for large-scale systems. In this new loss function, a small constant $c$ is added to balance both error terms so that the gradient descent works on both terms simultaneously. Our experiments show that by adding this new penalty term, the voltage phase angle estimation error is reduced from $12\%$ to $1.3\%$ in 30-bus system. 
\vspace{-0.1cm}
\begin{equation}
loss(\mathbf{z}, \mathbf{x})= \frac{1}{N}\sum_{i=1}^{N}(\mathbf{z} - h(\mathbf{x}))^T\mathbf{W}(\mathbf{z} - h(\mathbf{x})) + c\frac{1}{N}\sum_{i=1}^{N}(\mathbf{x} - \mathbf{\hat{x}})^2 \label{EQ4}
\end{equation}
\vspace{-0.1cm}

After the ANN models are trained, the testing data is used to evaluate their performance. A good SE model should preserve two properties: (1) provide accurate SE irrespective of the noise in the measurements; (2) bad-data alarms not triggered by regular measurement noise. Accordingly, we evaluate the estimation accuracy of the ANNs by \textit{maximum absolute relative error} (MARE) between the true and the estimated values. An estimation is considered accurate if MAREs of the voltage magnitude and the voltage phase angle do not exceed $1\%$ and $5\%$, respectively. Table\ref{tab3} summarizes the performance evaluation results based on a significant level $\alpha=0.01$ for the trained ANN models. It is clear from these tables that the proposed ANN models are able to estimate AC states accurately, and have low false alarm rate for bad data under regular measurement noises. 
\vspace{-0.2cm}
\begin{table}[ht]
\caption{ANN SE Model Evaluation}
\begin{center}
\begin{tabular}{c c c c c c}
\hline
\textbf{Test System}& \textbf{ MARE($|V|$)} & \textbf{Accuracy($|V|$)(\%)} & \textbf{ MARE($\theta$)}& \textbf{Accuracy($\theta$)(\%)}& \textbf{Bad Data(\%)}   \\
\hline
\textbf{9-bus}& $2.4\times 10^{-5}$ & $100$ & $1.6\times 10^{-2}$ & $96$  & $0$\\
\textbf{14-bus}& $5.6\times 10^{-5}$ & $100$ & $1.6\times 10^{-2}$ & $99$  & $3$\\
\textbf{30-bus}& $6.5\times 10^{-5}$ & $100$ & $1.3\times 10^{-2}$ & $98$ & $5$\\
\hline
\end{tabular}
\label{tab3}
\end{center}
\end{table}
\vspace{-0.6cm}
\section{Adversarial Model and Attack Formulation}

\vspace{-0.2cm}
\subsection{Adversarial Model}
\vspace{-0.3cm}
The goal of the attacker is to launch a FDI attack, in which the attacker aims to decide and inject a manipulated measurement vector into the measurement under given resource and meter accessibility constraints, such that the injection can maximize SE error while remaining stealthy. 

The attacker is assumed to have full knowledge of the topology and configuration of the power grid, such as the nodal admittance matrix. Such information could be accessed or estimated from public database or historical records. In addition, the attacker is also assumed to know everything about the ANN SE model, including the architecture and the parameters. These information could be obtained by the attacker either through breaking into the information system of the power grid (similar to the 2015 Ukraine case) or through training a shadow ANN that mimics the real ANN SE model on a substitute data set. The attacker is also assumed to know the threshold of the bad data detector. Although these assumptions render a strong attacker that may not always represent the practical cases, it enables us to evaluate the robustness and vulnerabilities of the ANN SE models under the worst-case scenario, providing an upper bound on the impact of FDI attacks on ANN AC SE. 

In addition to the bad data detection threshold, the adversary is also facing other constraints, including the set of meters she has access to, the maximum number of meters she can compromise, and the maximum amount of errors she can inject into a true measurement to avoid being detected.

Note that in this paper we only consider the FDI attacks that happen during the operational phase of the ANN SE. In other words, the adversary is only able to tamper the measurement inputs after the ANN model is trained. It is not allowed to perturb either the training data or the trained model. The investigation of training data and model pollution is out of the scope of this paper and will be studied in our future work. 
\vspace{-0.3cm}
\subsection{Attack Formulation}
Let $\textbf{z}_a$ be the measurement vector in the presence of FDI attack, then $\mathbf{z}_a$ can be described as:
\begin{equation}
\mathbf{z_a} = \mathbf{z} + \mathbf{a} = h(\mathbf{x}) + \mathbf{a}, 
\end{equation}
where $\mathbf{a}$ is a $N_m$-dimension non-zero attack vector. 
Given the input of manipulated measurement $\mathbf{z_a}$, the output by the ANN SE $f$ is as follows:
\begin{equation}
\mathbf{\hat{x}_{a}} = f(\mathbf{z_a}) = f(\mathbf{z + a})
\end{equation}
According to Eq.(\ref{EQ2}), an adversary intending to elude from bad data detection must satisfy the following condition:
\begin{equation}
J(\mathbf{\hat{x}_a})=(\mathbf{z_a} - h(\mathbf{\hat{x}_a}))^T\mathbf{W}(\mathbf{z_a} - h(\mathbf{\hat{x}_a})) \leq \tau
\end{equation}
The error injected to SE hence can be calculated by:
\begin{equation}
\mathbf{\hat{x}_{a}} - \hat{\mathbf{x}} = f(\mathbf{z_a}) - f({\mathbf{z}}). 
\end{equation}

With the above notations, the problem of finding the best adversarial injection $\mathbf{a}$ for a given measurement $\mathbf{z}$ can be formulated as a constrained optimization:
\begin{equation}
\begin{aligned}
&\underset{\mathbf{a}}{\text{maximize}}
& & \|\mathbf{\hat{x}_a} - \mathbf{\hat{x}}\|_p \\
& \text{subject to}
& & (\mathbf{z} - h(\mathbf{\hat{x}_a}))^T\mathbf{W}(\mathbf{z_a} - h(\mathbf{\hat{x}_a})) < \tau,\\
& & & \|\mathbf{a} \|_0 \leq L,\\
& & & a_i^l \leq a_i \leq a_i^u, i=1,...,N_m,\\
& & & z_i^{min} \leq z_{a_i} \leq z_i^{max}, i=1,...,N_m,\\
\end{aligned} \label{EQ9}
\end{equation}
where $L$ is the maximum number of meters the attacker can compromise, $[a_i^l, a_i^u]$ provides limits of modification to each compromised meter, and $[z_i^{min}, z_i^{max}]$ denotes the valid range for each measurement, ensuring the manipulated measurement to still be within the power range permitted on that particular unit. The strength of measurement modification depends on the attacker's resource and meter accessibility constraints, which have not been considered in previous work. In our work, by limiting the measurement manipulation to a subset of meters, we are able to prevent from injecting excessive errors, which can be easily detected by univariate analysis. In addition, if the adversary can locate high precision meters, she can avoid injecting too much errors into those meters and instead allocate resource to other meters to improve the overall attack outcomes.

The objective function in the optimization Eq.(\ref{EQ9}) requires a distance metric $\|\cdot\|_p$ to quantify attack impact. This distance metric should be carefully defined to reflect the severity of physical impact on the power grid caused by the SE error. In reality, the voltage magnitudes in the state are always limited in a tight range to ensure stable electricity supply, while the voltage phase angles could vary in a relatively large range, and hence an erroneous estimation of the latter may seriously affect the consistent operation of the power grid, but cannot be easily detected. Therefore, we define the adversary's objective function as the maximum change to the voltage phase angles $\mathbf{\theta}$:
 \begin{equation}
 \|\mathbf{\hat{x}_a} - \mathbf{\hat{x}}\|_\infty  = \max(|\hat{\theta}_{a_1} - \hat{\theta}_1|, ...,|\hat{\theta}_{a_n} - \hat{\theta}_n|)
 \end{equation}

\section{Attack Methodology}
\vspace{-0.2cm}
\subsection{Solving the Proposed Attack with DE}
\vspace{-0.1cm}
As a population based stochastic optimization algorithm, DE algorithm was first proposed in 1996 by Rainer et al. \cite{storn1997differential}. The population is randomly initialized within the variable bounds. In each generation, a mutant vector is produced by adding a target vector (father) with a weighted difference of other two randomly chosen variables. Then a crossover parameter mixes father and mutant vector to form a candidate solution  (child). A comparison is drawn between father and child, whichever that is better will enter the next generation. 

We follow \cite{Su2017} to encode our measurement attack vector into an array, which contains a fixed number of perturbations, and each perturbation holds two values: the compromised meter index and the amount to inject to that meter. The use of DE and the encoding has the following three advantages: \textbf{1. Higher probability of finding global optimum} - In each generation, the diversity introduced by mutation and crossover operations ensures the solution not stuck in local optimum, and thus leads to a higher probability of finding global optimum. \textbf{2. Adaptability for multiple attacks} - DE can adapt to different attack scenarios based on our encoding method. By  DE can search for both meter indices and injection amount or only search for injection amount to these specified meters, by specifying the number of meters to compromise or fixing the meter indices. \textbf{3. Parallelizibility to shorten attack time} -  As the smart grid scale increases, generating one attack vector may take from seconds to minutes. An attacker must finish attack vector generation and injection before next SE takes place. As it is based on a vector population, DE is parallelization friendly, so as to significantly expedite the computation for the attack vector.

Next, we present how we adapt DE algorithm to our proposed attack. The pseudo code for the proposed attack using DE is presented in Algorithm 1:
\begin{itemize}
\item \textbf{Deal with duplicate meter indices} - Instead of outputting the exact meter value, we select to output the injection vector to narrow down search space. We use two approaches to ensure the uniqueness of meter indices in the solution. First, we generate meter indices without replacement in population initialization. Second, we add a filter in the crossover operation. This filter keeps the meter indices unchanged if the newly selected meter index is repetitive with previous meter indices. 

\item \textbf{Ensure the measurement after injection is within range} - A valid measurement reading must satisfy $z_i^{min} \leq z_i + a_i \leq z_i^{max}$ for all $i$, where $z_i^{min}$ and $z_i^{max}$ are lower and upper limit power permitted on $z_i$. We use an intuitive approach by replace $\mathbf{z_a} = \mathbf{z} + \mathbf{a}$ with $\mathbf{z_a} = min(max(\mathbf{z_a}, \mathbf{z}^{min}), \mathbf{z}^{max})$, where the min and max are element-wise operations. 
\item \textbf{Deal with the overall constraint} - In addressing constraints with DE, using a penalty function has been the most popular approach. However, they do not always yield satisfactory solutions since the appropriate multiplier for the penalty term is difficult to choose and the objective function may be distorted by the penalty term. Therefore, we use a heuristic constraint handling method \cite{deb2000efficient}. A pair-wise comparison is performed between fathers and children in order to differentiate better solutions from population. The three criteria of the pair-wise comparison are as the following: 1. If both vectors are feasible, the one with the best objective function value is preferred. 2. If one vector is feasible and the other one is not, the feasible one is preferred. 3. If both two vectors are infeasible, the one with the smaller constraint violation is preferred. The above comparisons handle constraint in two steps: first, the comparison among feasible and infeasible solutions provides a search direction towards the feasible region; then, the crossover and mutation operations keep the search near the global optimum, while maintaining the diversity among feasible solutions.
\end{itemize}

\begin{algorithm}[ht]
\caption{DE attack}
\begin{algorithmic}[1]
\renewcommand{\algorithmicrequire}{\textbf{Input:}} 
\renewcommand{\algorithmicensure}{\textbf{Output:}}
\REQUIRE{measurement $\mathbf{z}$, $GEN_{MAX}$\{maximum number of generations\}, $N$\{population size\}, $f$\{objective function\}, g\{constraint function\}}, $CR$\{crossover rate\}
\ENSURE {injection vector $\textbf{a}$}
\STATE $g = 0$
\STATE Population initialization $\textbf{a}_{i, 0}$ for  $i=1,...,N$. Meter indices are randomly select without replacement and  injection amounts are randomly select within the univariate bound.
\STATE Evaluate the $f(\mathbf{a}_{i, g})$ and constraint violation $CV(\mathbf{a}_{i, g}) = max(g(\mathbf{a}_{i, g}), 0)$, for $i=1,...,N$
\FOR{$g = 1: MAX_{GEN}$}
    \FOR{$i = 1:N$}
        \STATE{Randomly select $r_1$ and $r_2$}
        \STATE{$j_{rand}=randint(1, N_m)$}
        \FOR{$j=1:D$}
            \IF{($rand_j[0,1)<CR$ \OR $j=j_{rand}$) \AND the meter index not repetitive with previous meter indices}
                \STATE{$u^j_{i, g+1} = x^j_{best, G} + F(x^j_{r_1, g} - x^j_{r_2, g})$}
            \ELSE
                \STATE{$u^j_{i, g+1} = x^j_{i, G}$}
            \ENDIF
        \ENDFOR
        \STATE{Evaluate $f(\mathbf{u}_{i, g+1})$ and $CV(\mathbf{u}_{i, g+1})$}
        \STATE{Update the population if the child $\mathbf{u}_{i, g+1}$ is better than the father $\mathbf{x}_{i,g}$ by the above three criteria}
    \ENDFOR
\ENDFOR
\end{algorithmic}
\end{algorithm}
\vspace{-0.5cm}
\subsection{Solving the Proposed Attack with SLSQP}
\vspace{-0.1cm}
In some gradient based attack algorithms in image classification(\cite{szegedy2013intriguing}, \cite{Carlini2017a}), the logistic function is added to the objective function as a penalty term and the parameter for the penalty term is chosen by line search. These algorithms aim to find a feasible feasible solution, not the optimal one. Therefore, we use a conventional optimization algorithm (SLSQP) \cite{software}. SLSQP is a variation on the SQP algorithm for non-linearly constrained gradient-based optimization. In our SLSQP attack, we encode the solution to a $N_m$-dimension vector, in which the $ith$ element denotes the injection amount to the $ith$ meter. This encoding allows the attacker to generate attack vectors with a set of specified meters by placing upper and lower bounds to corresponding elements in the attack vector. 
To solve the proposed optimization problem, we first construct the Lagrangian function:
\begin{equation}
\mathcal{L}(\mathbf{a}, \lambda) = f(\mathbf{a}) + \lambda\cdot g(\mathbf{a}),
\end{equation}
\vspace{-0.1cm}
where 
\vspace{-0.1cm}
\begin{equation}
\left \{ \begin{array}{l}
f(\mathbf{a}) = \|\mathbf{\hat{x_a}} - \mathbf{\hat{x}}\|_\infty \\
g(\mathbf{a}) =  (\mathbf{z} - h(\mathbf{\hat{x_a}}))^T\mathbf{W}(\mathbf{z_a} - h(\mathbf{\hat{x_a}})) < \tau\\
\end{array} \right.
\end{equation}
\vspace{-0.1cm}
In each iteration $k$, the above problem can be solved by transferring to a linear least square sub-problem in the following form:
\vspace{-0.1cm}
\begin{equation}
\begin{aligned}
&\max_{\mathbf{d}} 
& & \|(\mathbf{D^k)}^{1/2}(\mathbf{L^k})^T\mathbf{d} +  ((\mathbf{D^k})^{-1/2}(\mathbf{L^k})^{-1}\nabla(\mathbf{a}^k)\|\\
& \text{subject to}
& & \nabla g(\mathbf{a}^k)\mathbf{d} + g(\mathbf{a}^k) \geq 0
\end{aligned}
\end{equation}
\vspace{-0.1cm}
where $L^k D^k (L^k)^T$ is a stable factorization of the chosen search direction $\nabla_{zz}^2 \mathbf{L}(\mathbf{z}, \lambda)$ and is updated by BFGS method. By solving the QP sub-problem for each iteration, we can get the value of $\mathbf{d}^k$, i.e., the update direction for $\mathbf{z}^k$:
\vspace{-0.1cm}
\begin{equation}
\mathbf{z}^{k+1} = \mathbf{z}^k + \alpha \mathbf{d}^k
\end{equation}
\vspace{-0.1cm}
where $\alpha$ is the step size, which is determined by solving an additional optimization. The step size $\psi(\alpha):=\phi(\mathbf{a}^k+\alpha d^k)$ with $\mathbf{x}^k$ and $d^k$ are fixed, can be obtained by a minimization:
\vspace{-0.1cm}
\begin{equation}
\phi(\mathbf{a}^k; r) := f(\mathbf{a^k}) + max(r\cdot g(\mathbf{a}), 0)
\end{equation}
\vspace{-0.1cm}
with  $r$ being updated by:
\vspace{-0.1cm}
\begin{equation}
r^{k+1}:=max(\frac{1}{2}(r^k + |\lambda|, |\lambda|))
\end{equation}
\section{Attack Evaluation}
\vspace{-0.3cm}
Here, both FDI attacks are evaluated on IEEE 9-bus, 14-bus, and 30-bus test systems. The simulateion is done in Python, using package \textit{TensorFlow} and \textit{SciPy}, on a computer with a 3.5 GHz CPU and a 16 GB memory. 

Depending on the attacker's capability and practical constraints, the attacker can launch attack under different scenarios. Inspired by \cite{liu2011false}, we construct two attack scenarios to facilitate the evaluation: (1)\textbf{Any $k$ meter attacks.} The attacker can access all meters, but the number of meters to compromise is limited by $k$. In this scenario, the attacker can wisely allocate the limited resources, by selecting meters and injection amounts that will maximize her attack impact. (2)\textbf{Specific $k$ meter attacks.} The attacker has the access to $k$ specific meters. For example, the attacker may only access the a set of meters in a small region. She needs to determine injection amount to maximize attack impact.

We perform the experiment as follows. To fairly compare attack performance on different test systems, we choose the percentage of meters being compromised, $R$,  to be $5\%$, $10\%$ and $20\%$. For each $R$, we explore the attack performance under different error injection levels: $2\%$, $5\%$ and $10\%$. Each experiment runs on 1000 measurement instances, and is repeated for 10 times to reduce randomness.

We consider four metrics throughout evaluating the effectiveness of the attacks. We measure the MAE and MARE of the error injected to voltage phase angles. We also report the success probability, where success is defined as the attack produce more than $1\%$ or $5\%$ MARE to voltage magnitude or phase angle, respectively. Moreover, since the smart grid is assumed to be a quasi-static system and the states change slowly over time, we want to investigate if the time allows an adversary to mount an FDI attack to smart grid.
\vspace{-0.3cm}
\subsection{Any $k$ Meter Attack}
\vspace{-0.3cm}
Under this scenario, the attacker can access all meters and has freedom to choose any $k$ meters to compromise. The way we encode the attack vector in DE enables the search for better meters in every generation. In contrast, SLSQP only allows us to put constraint on specific meter indices. Therefore, only DE can be used to find attack vectors in any $k$ meter attack. 

\begin{figure}[htpb]
\centering
\subfigure[Voltage Phase Angle]{
\includegraphics[width=0.28\linewidth]{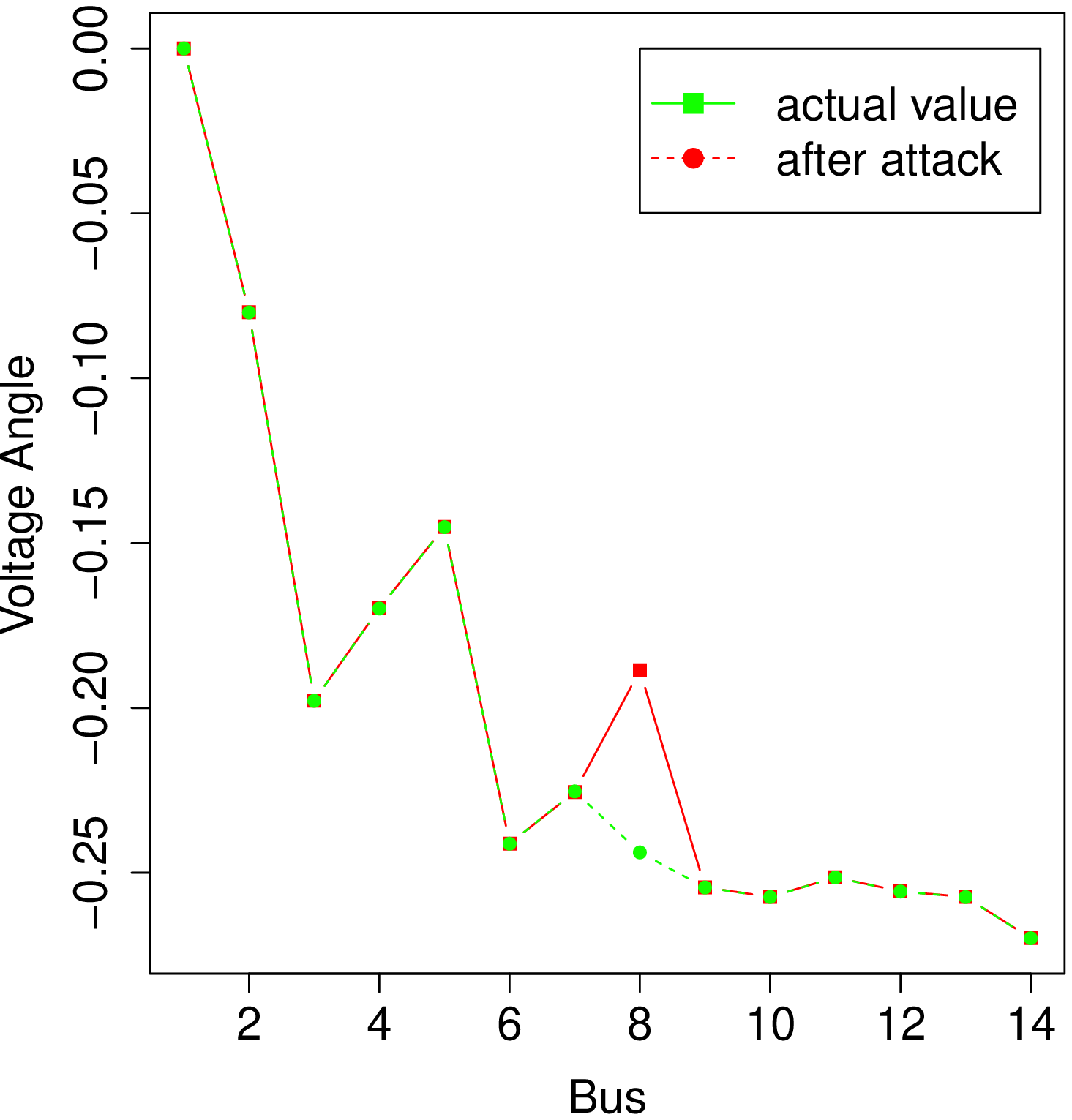}
}
\quad
\subfigure[CDF of Absolute Error]{
\includegraphics[width=0.28\linewidth]{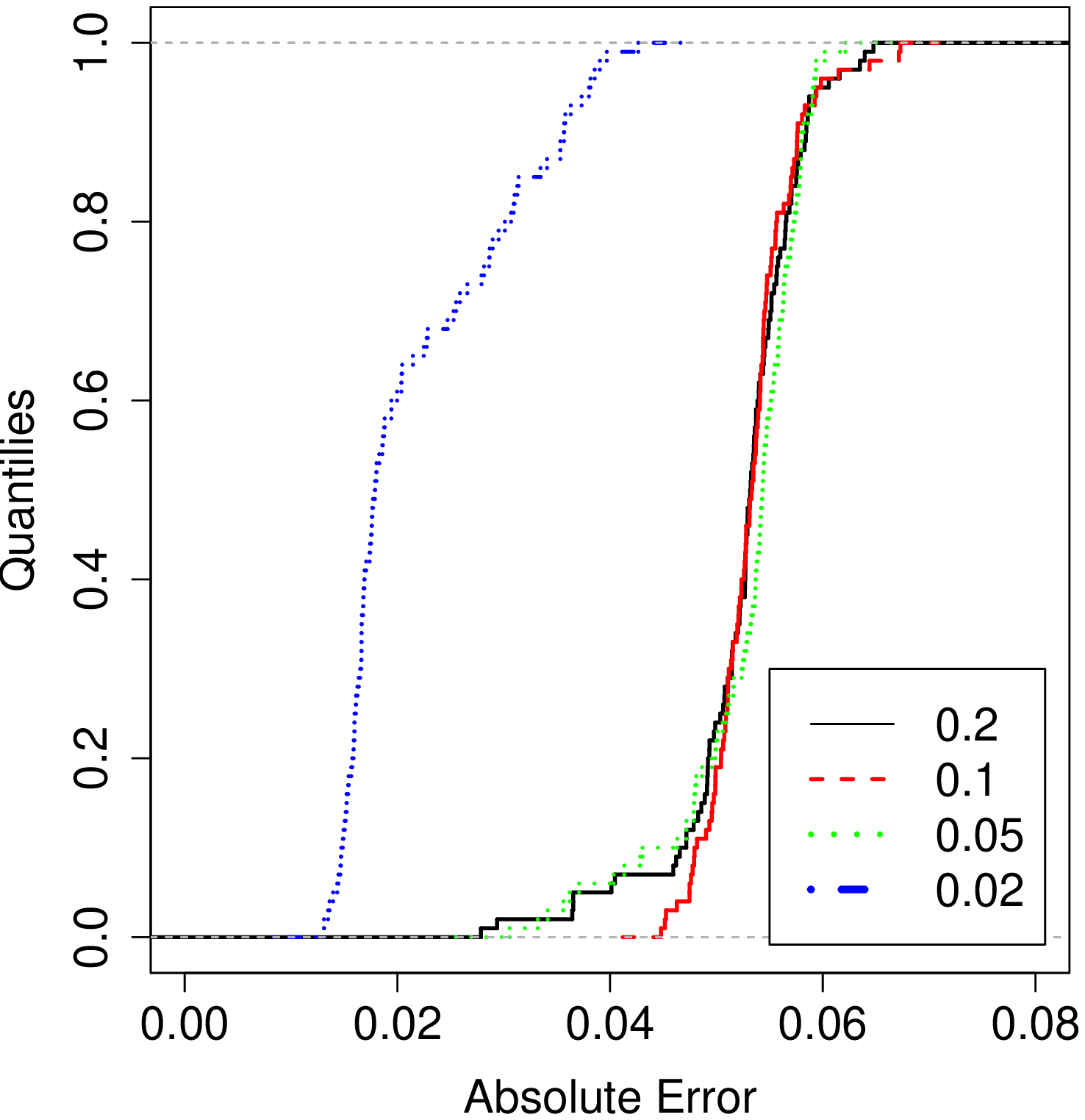}
}
\quad
\subfigure[CDF of Maximum Injection]{
\includegraphics[width=0.28\linewidth]{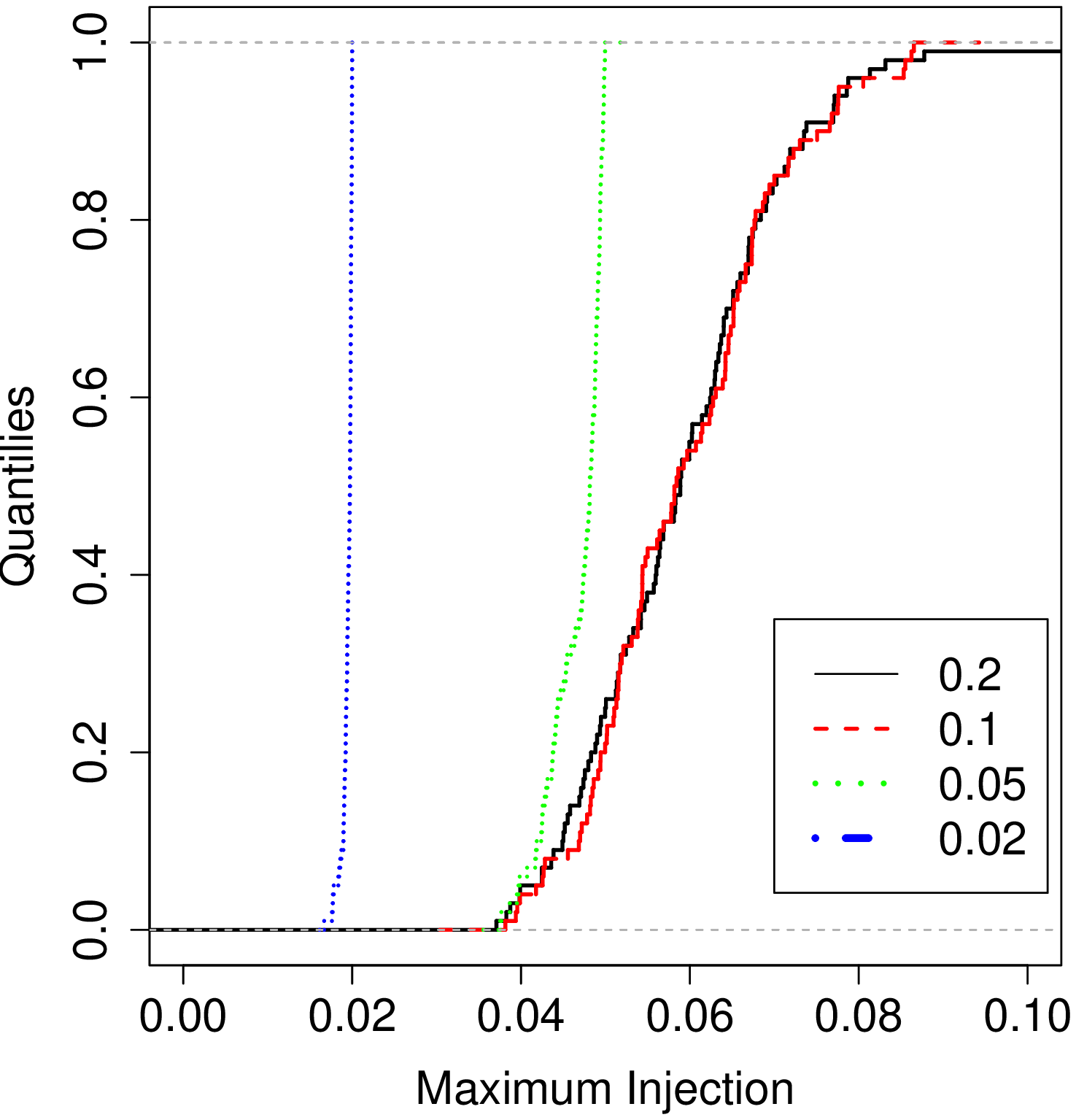}
}
\caption{An Example of a 5-meter Attack to 14-bus System }
\label{fig3}
\end{figure}

Our DE attack inject error to one of voltage phase angles while other values keep unchanged. In Figure~\ref{fig3} (b) and (c), for injection level $10\%$ and $20\%$, the maximum injections are concentrated around $5\%$ and seldom go beyond $10\%$, due to the overall constraint of bad data detection.
\begin{figure}[ht]
\centering
\subfigure[9-bus Relative Error]{
\includegraphics[width=0.28\linewidth,height=4cm]{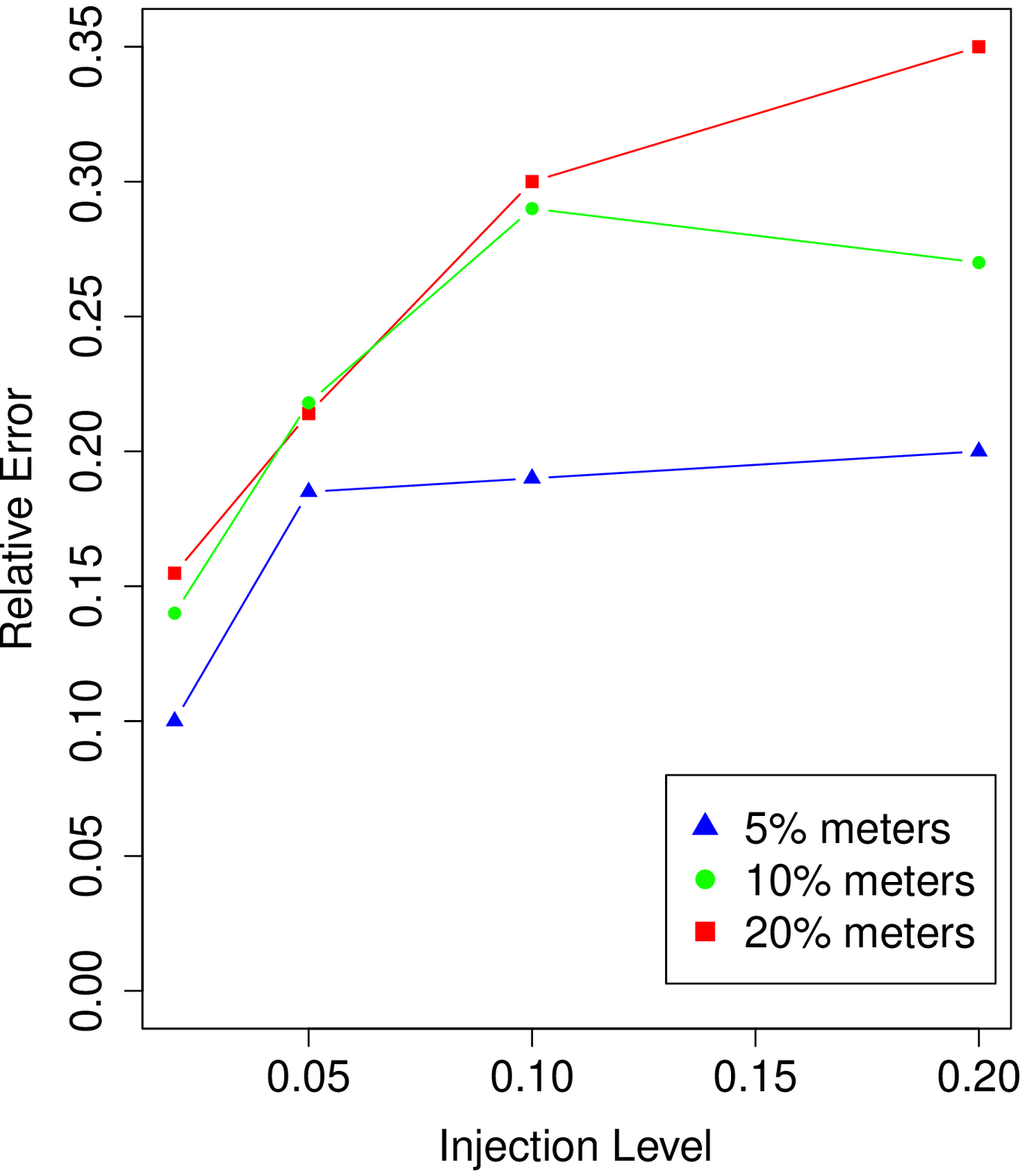}
}
\quad
\subfigure[14-bus Relative Error]{
\includegraphics[width=0.28\linewidth,height=4cm]{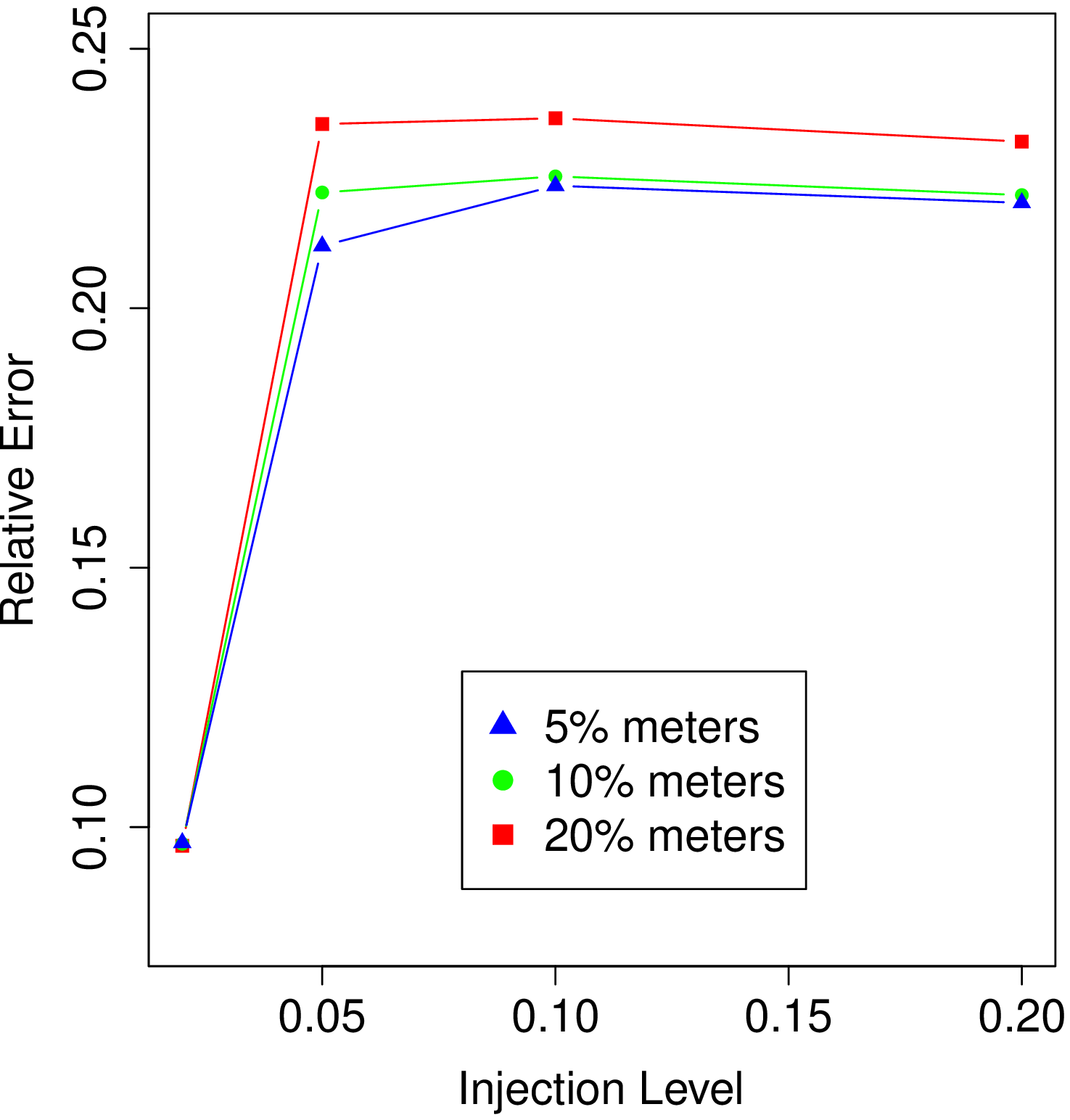}
}
\quad
\subfigure[30-bus Relative Error]{
\includegraphics[width=0.28\linewidth,height=4cm]{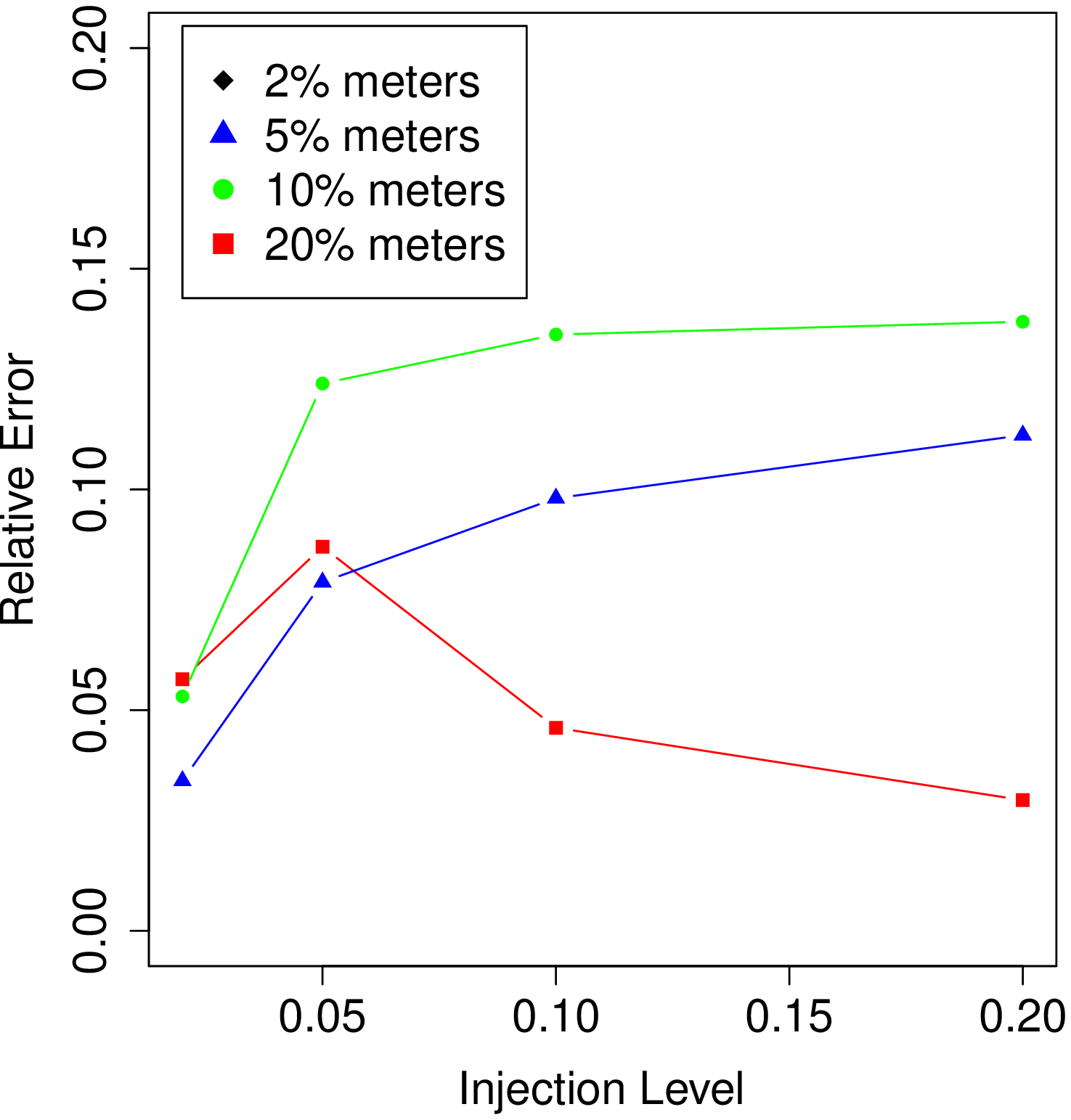}
}
\quad
\subfigure[9-bus Success Prob.]{
\includegraphics[width=0.28\linewidth,height=4cm]{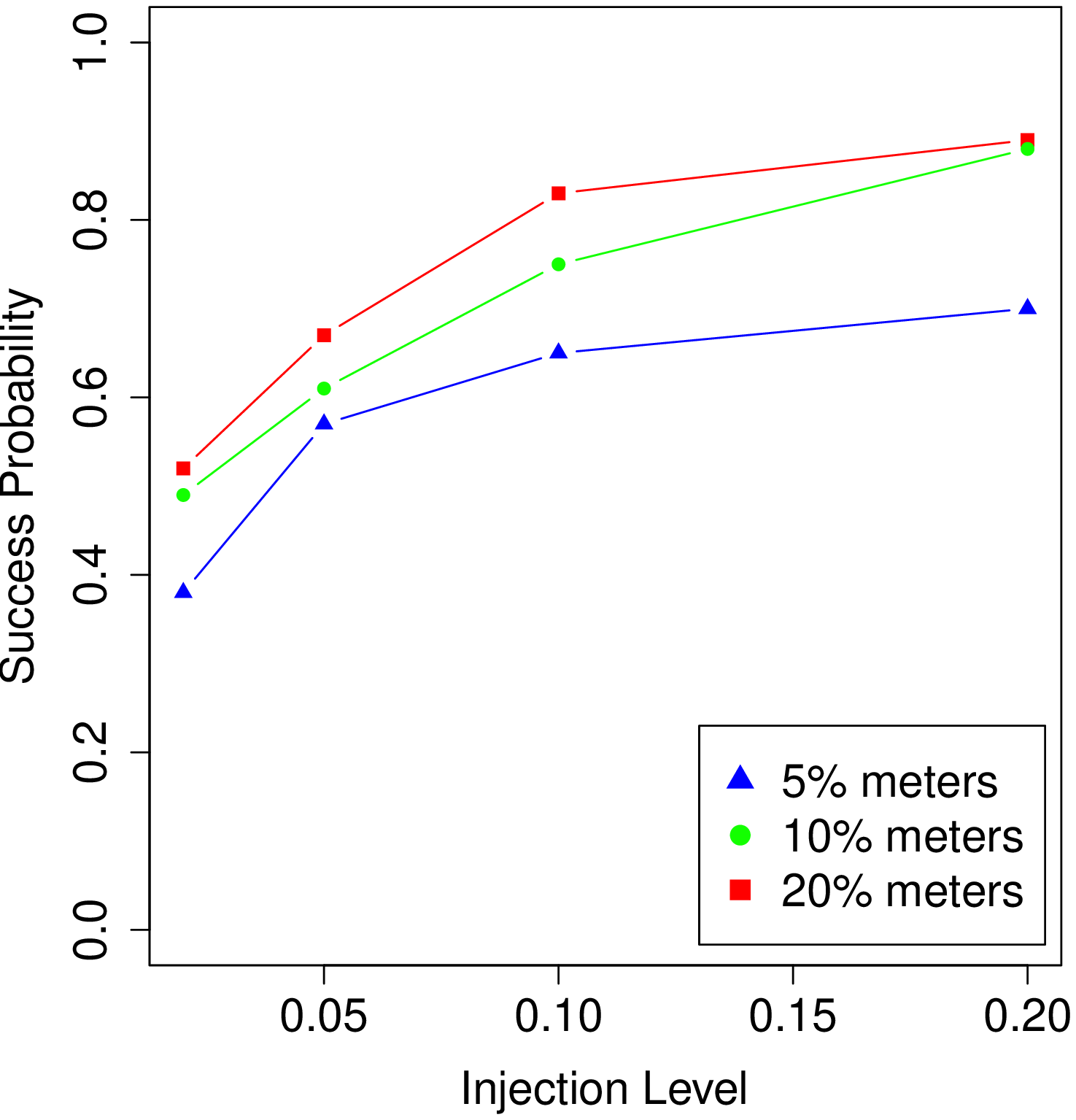}
}
\quad
\subfigure[14-bus Success Prob.]{
\includegraphics[width=0.28\linewidth,height=4cm]{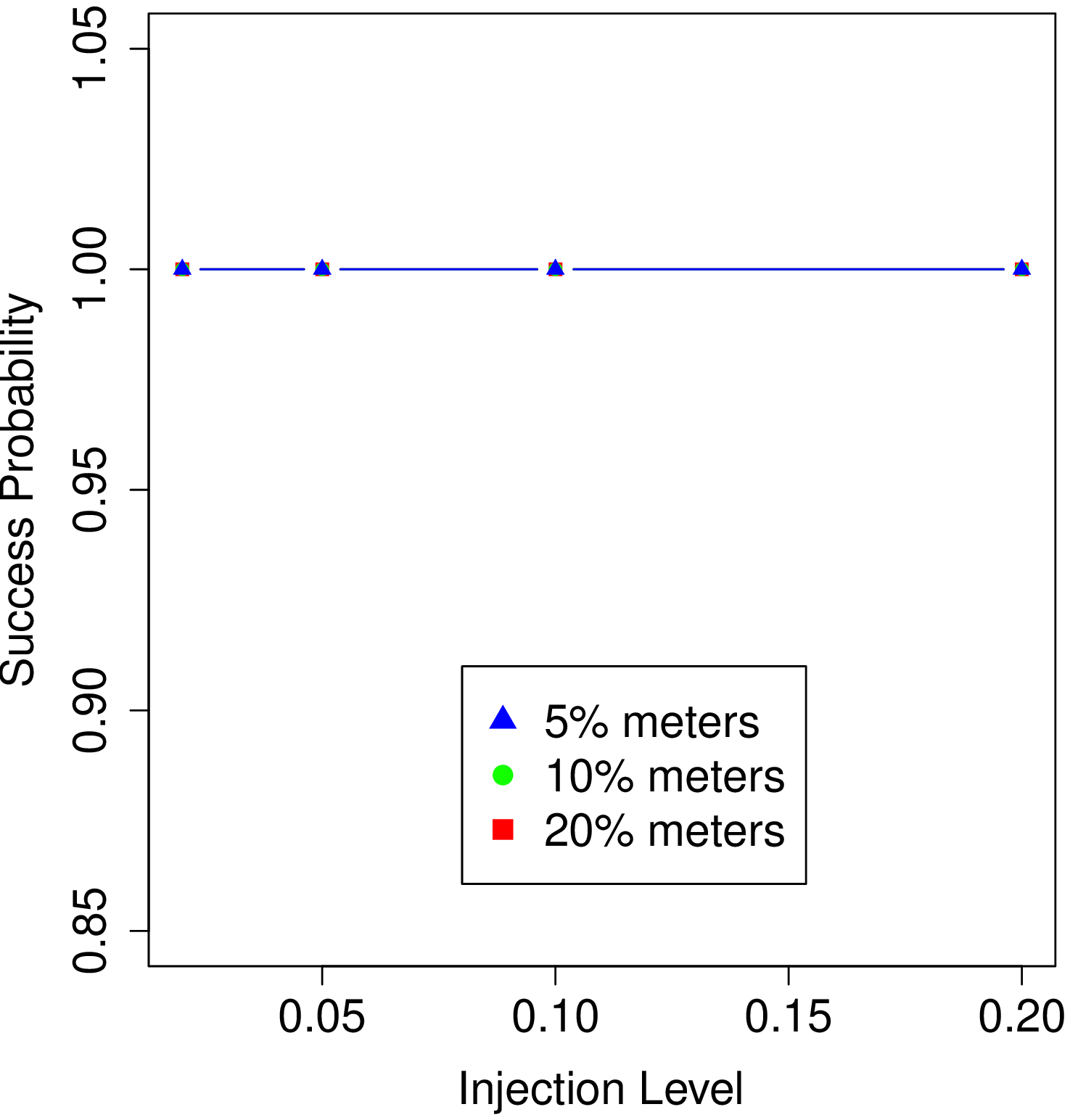}
}
\quad
\subfigure[30-bus Success Prob.]{
\includegraphics[width=0.28\linewidth,height=4cm]{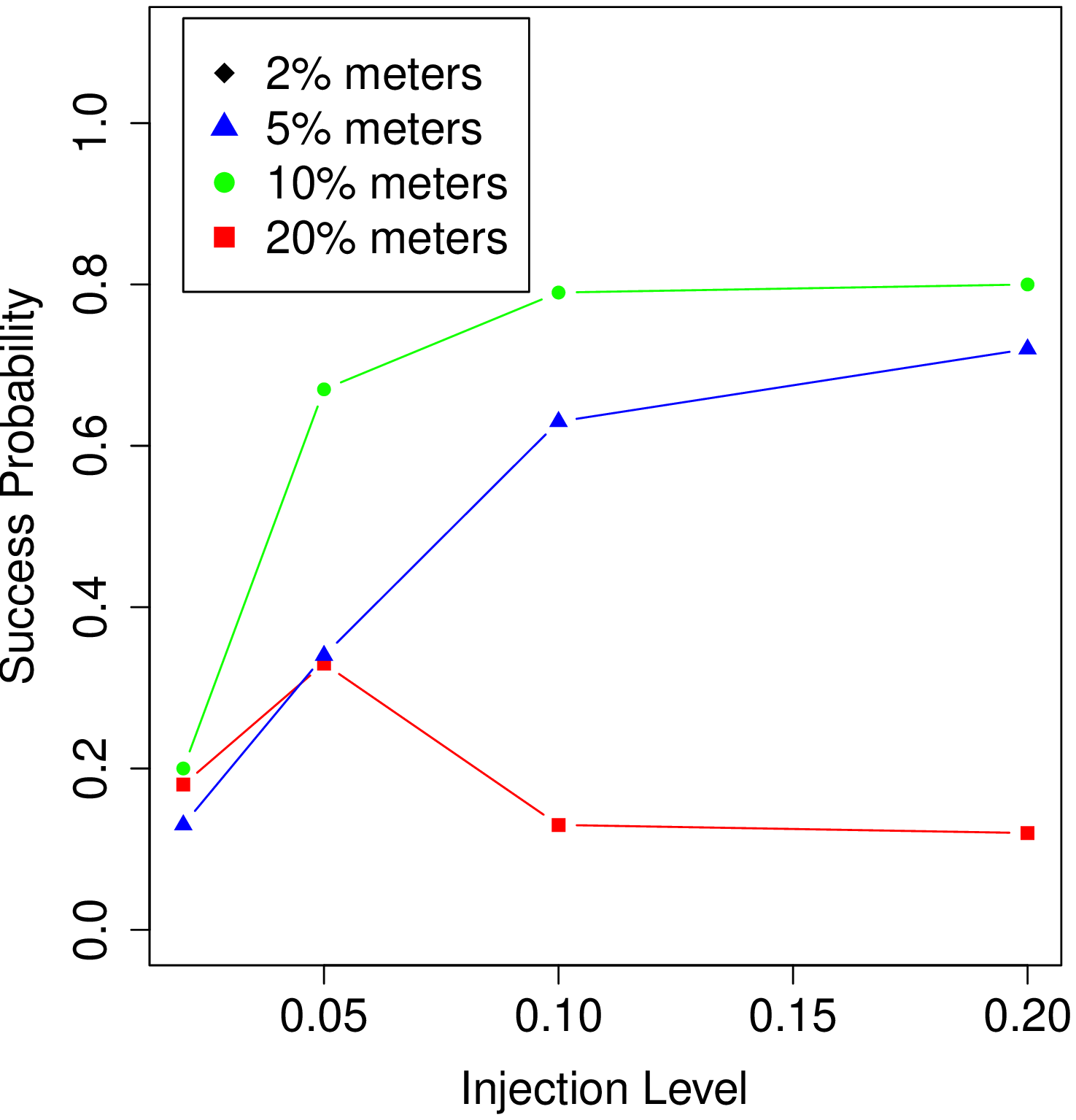}
}\quad
\caption{Relative Error (first row) and Success Prob. (second row) of Any $k$ Meter Attack with $N =400$ and $G_{MAX} =400$ }
\label{fig1}
\end{figure}

In general, the success probability and attack impact increase as the attacker controls more resource. The attack would succeed with larger probability ($80\%$ of simulation instances) by compromising $10\%$ of meters with injection level $10\%$. Especially for 14-bus system, the attack achieves $100\%$ success for any combination of $R$ and injection level (Figure \ref{fig1}). 

Interestingly, for 30-bus system, the impact of compromising $10\%$ of meters surpasses that of compromising $20\%$ of meters. Moreover, the performance of $20\%$ of meter compromised drops drastically as the injection level increases. A possible explanation for this might be that, as the expansion of search dimension and space, it would require more attempts to find a satisfactory solution.

\begin{figure}[ht]
\centering
\subfigure[9-bus]
{
    \centering
    \includegraphics[width=0.30\linewidth]{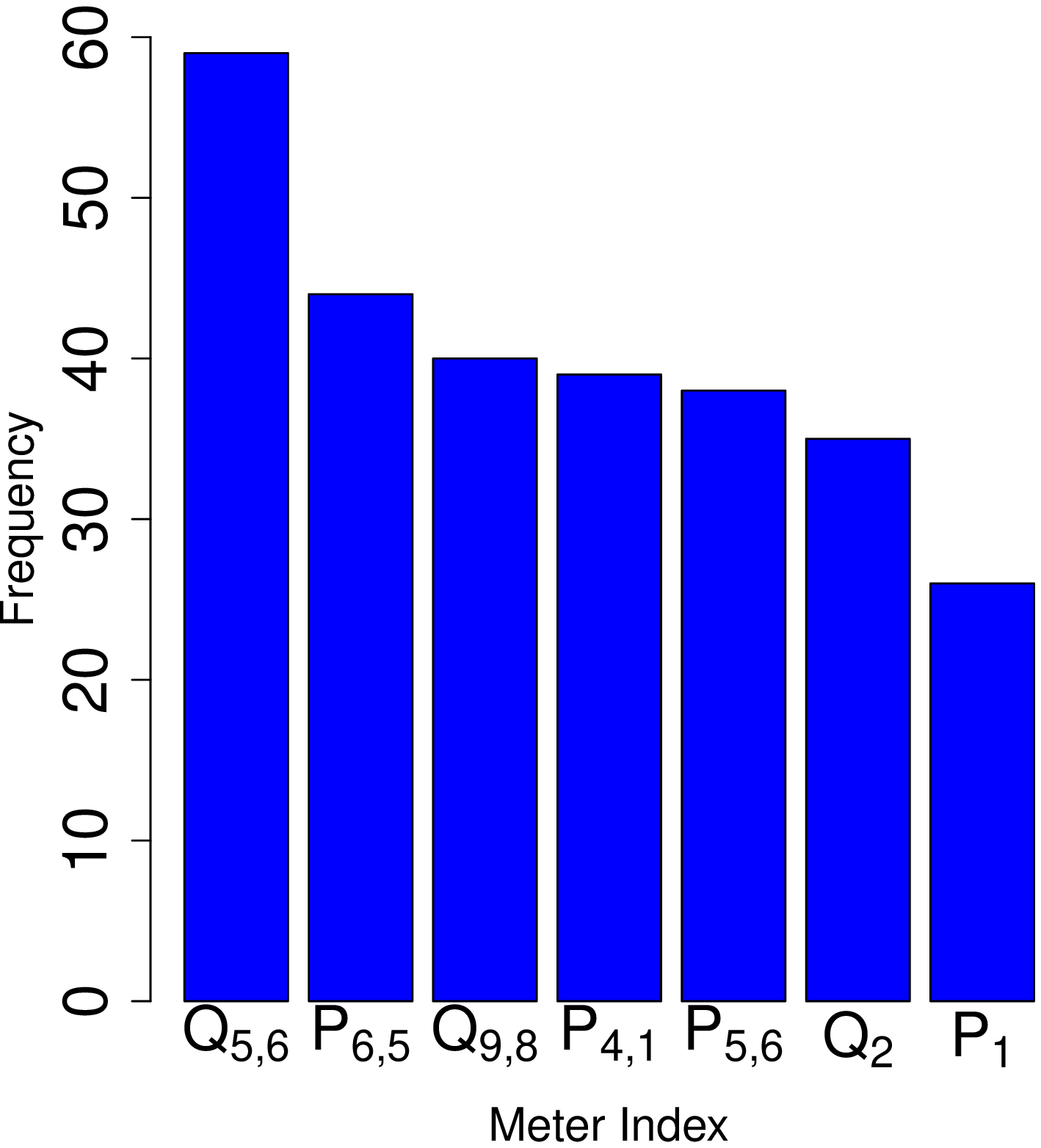}
}
\subfigure[14-bus]{
    \centering
    \includegraphics[width=0.30\linewidth]{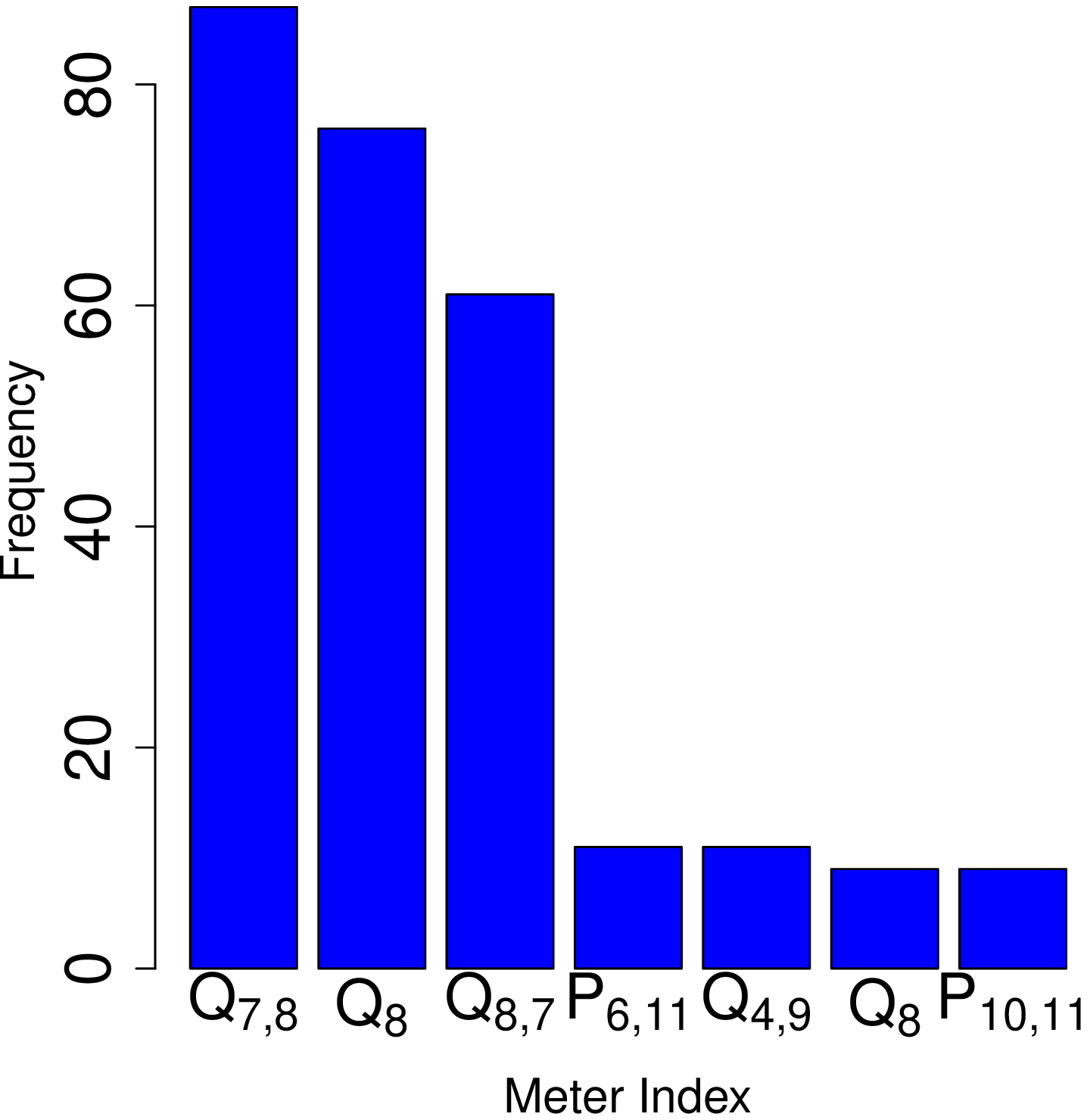}
}
\subfigure[30-bus]{
    \centering
    \includegraphics[width=0.30\linewidth]{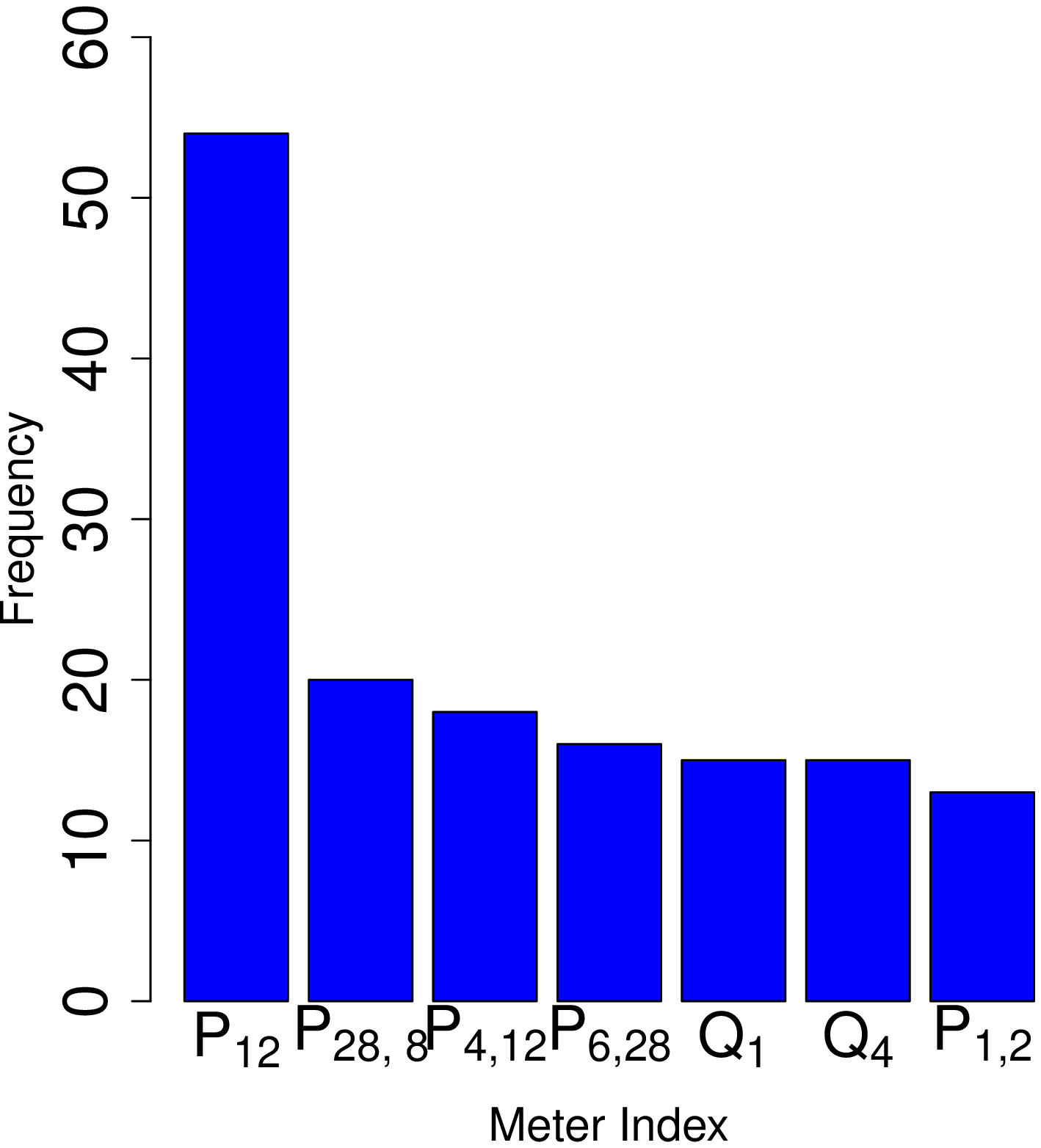}
}
\caption{Cumulative Frequency of Meters Presenting in Attack Vectors} 
\label{fig2}
\end{figure}
Figure~\ref{fig2} shows the first seven cumulative meter indices' frequency in the attack vectors. Injection to meters with higher frequency can introduce larger errors to state variables. Our DE attacks provide a practical way for systematically identifying key meters whose readings have a higher weight on the AC SE, and thus may guide the utility company to reach a more focused protection towards these key meters under resource and budget constraints.
\begin{table}[htpb] 
\caption{Average NFEs and Execution Time (in second) of Any $k$ Attacks}
\begin{center}
\renewcommand{\arraystretch}{1.2}
\begin{tabular}{*3c }
\hline
\textbf{Test System}& \textbf{NFEs} & \textbf{Time (s)}\\
\hline
9-bus & 500-1500 & 0.25-0.45 \\
14-bus & 500-3500 & 0.5-1.73\\
30-bus & 800-5600 & 1.5-2.7\\
\hline
\end{tabular}
\label{tab5}
\end{center}
\end{table}
\vspace{-0.3cm}
\subsection{Specific $k$ Meter Attack}
To explore the effect of population size and iteration number, we evaluate the average \textit{number of function evaluations} (NFEs) before delivering a successful attack or there is no significant change in the solution. The NFEs and corresponding running time are shown in Table \ref{tab5}. 

In this constrained scenario, the attacker is able to compromise specific $k$ meters due to physical location restrictions. DE and SLSQP are implemented and compared under this attack scenario. To search the injection amounts to specific $k$ meters, DE specifies the indices of the $k$ meters in population initialization and disables the mutation operation, while SLSQP only allows modifications to the $k$ meters in the attack vector. We randomly select $R$ to be $5\%$, $10\% $ and $20\%$ from test systems and perform the same set of experiments using both DE and gradient-based algorithm and compare their performance by the same metrics.

\begin{figure}[ht]
\centering
\subfigure[9-bus Relative Error]{
\includegraphics[width=0.28\linewidth, height=4cm]{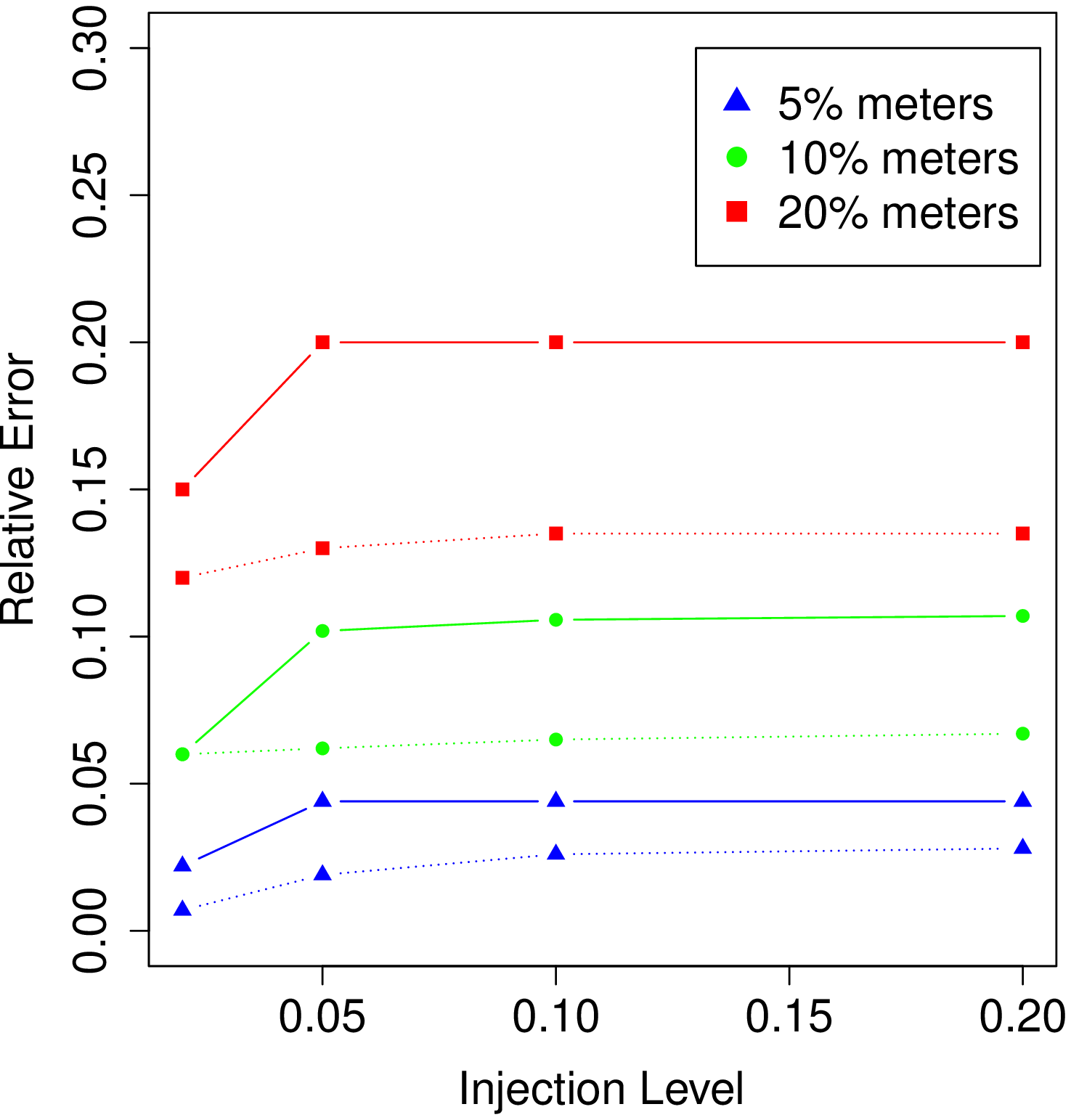}
}
\quad
\subfigure[14-bus Relative Error]{
\includegraphics[width=0.28\linewidth, height=4cm]{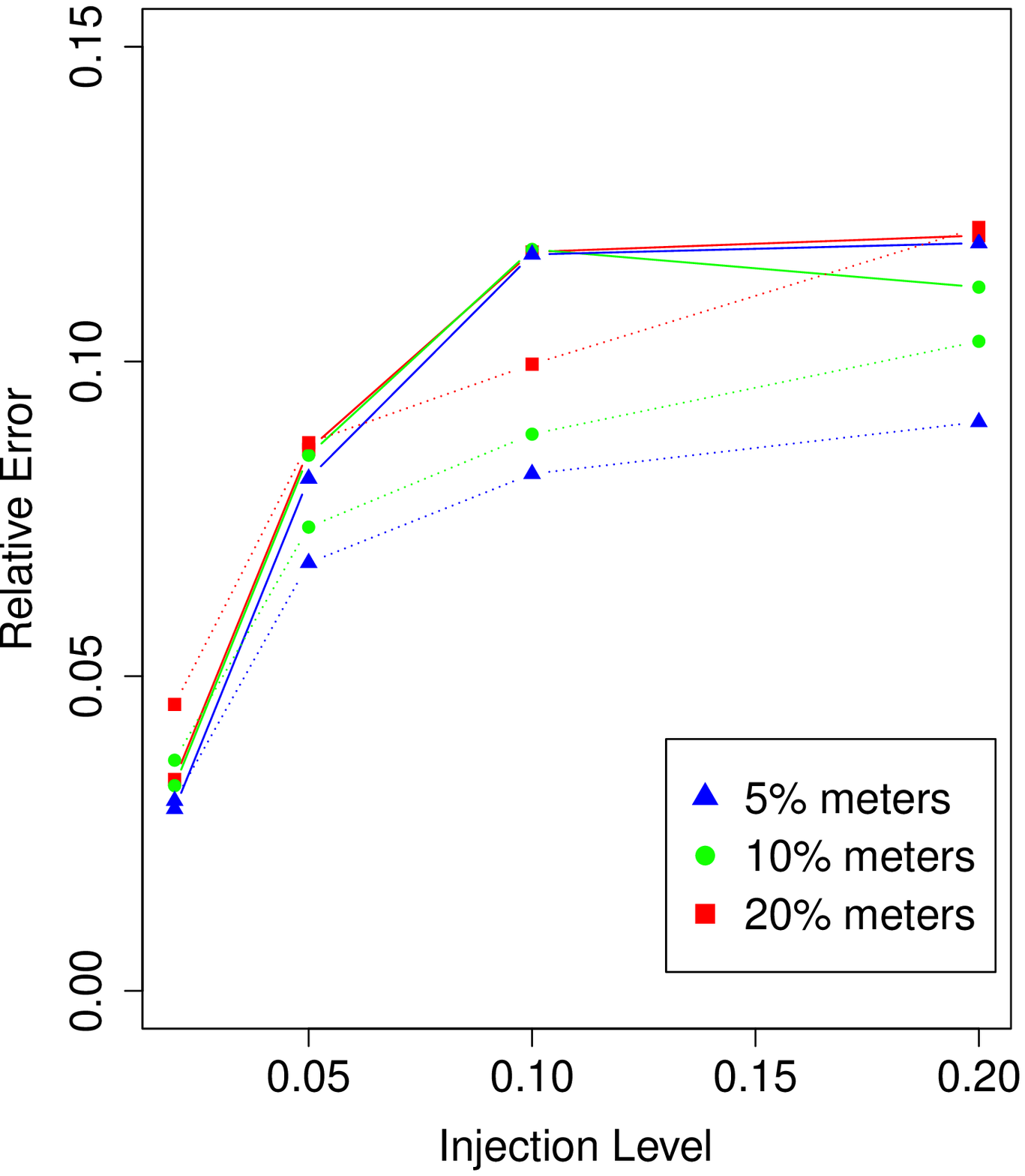}
}
\quad
\subfigure[30-bus Relative Error]{
\includegraphics[width=0.28\linewidth,height=4cm]{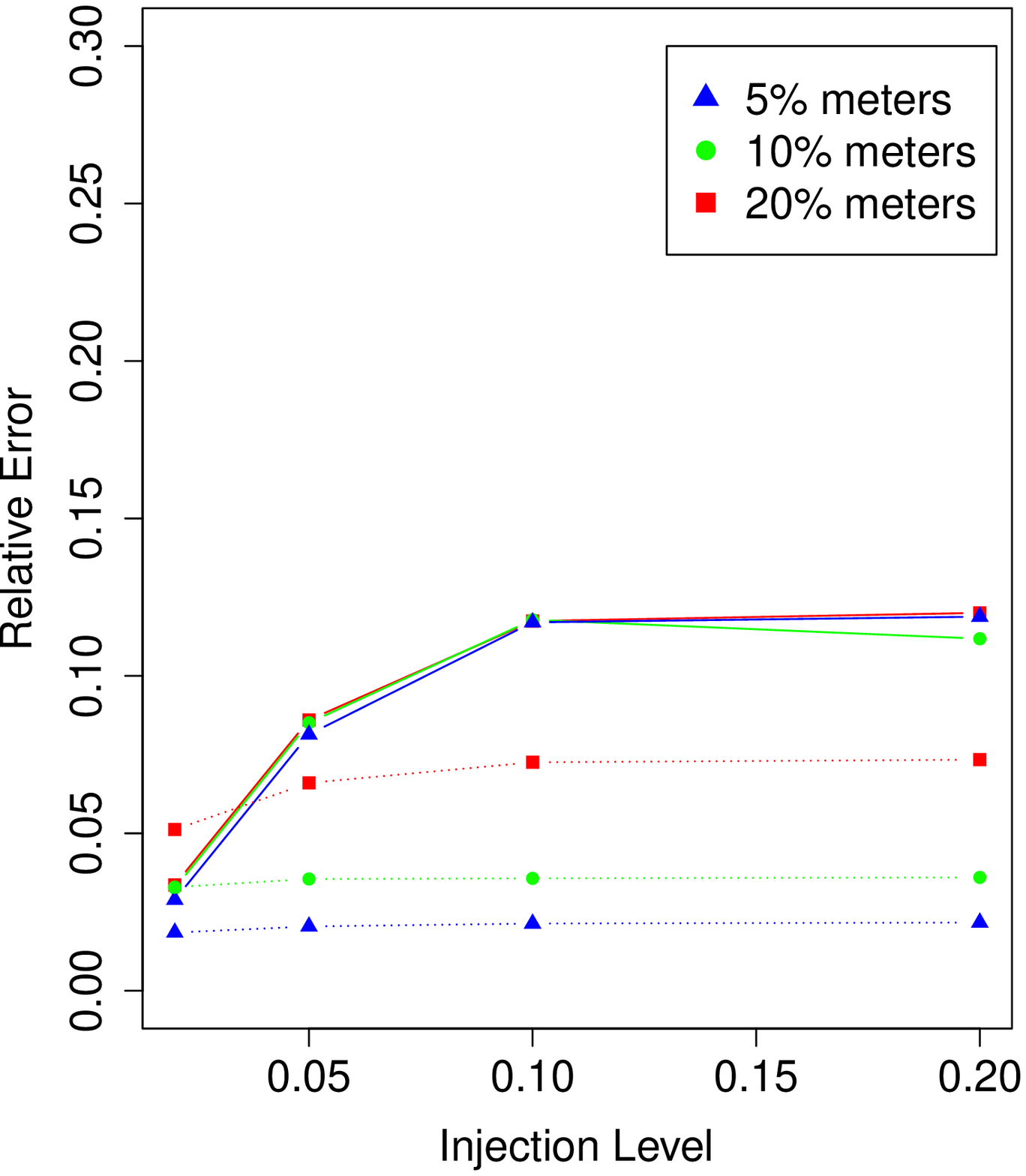}
}
\quad
\subfigure[9-bus Success Prob.]{
\includegraphics[width=0.28\linewidth,height=4cm]{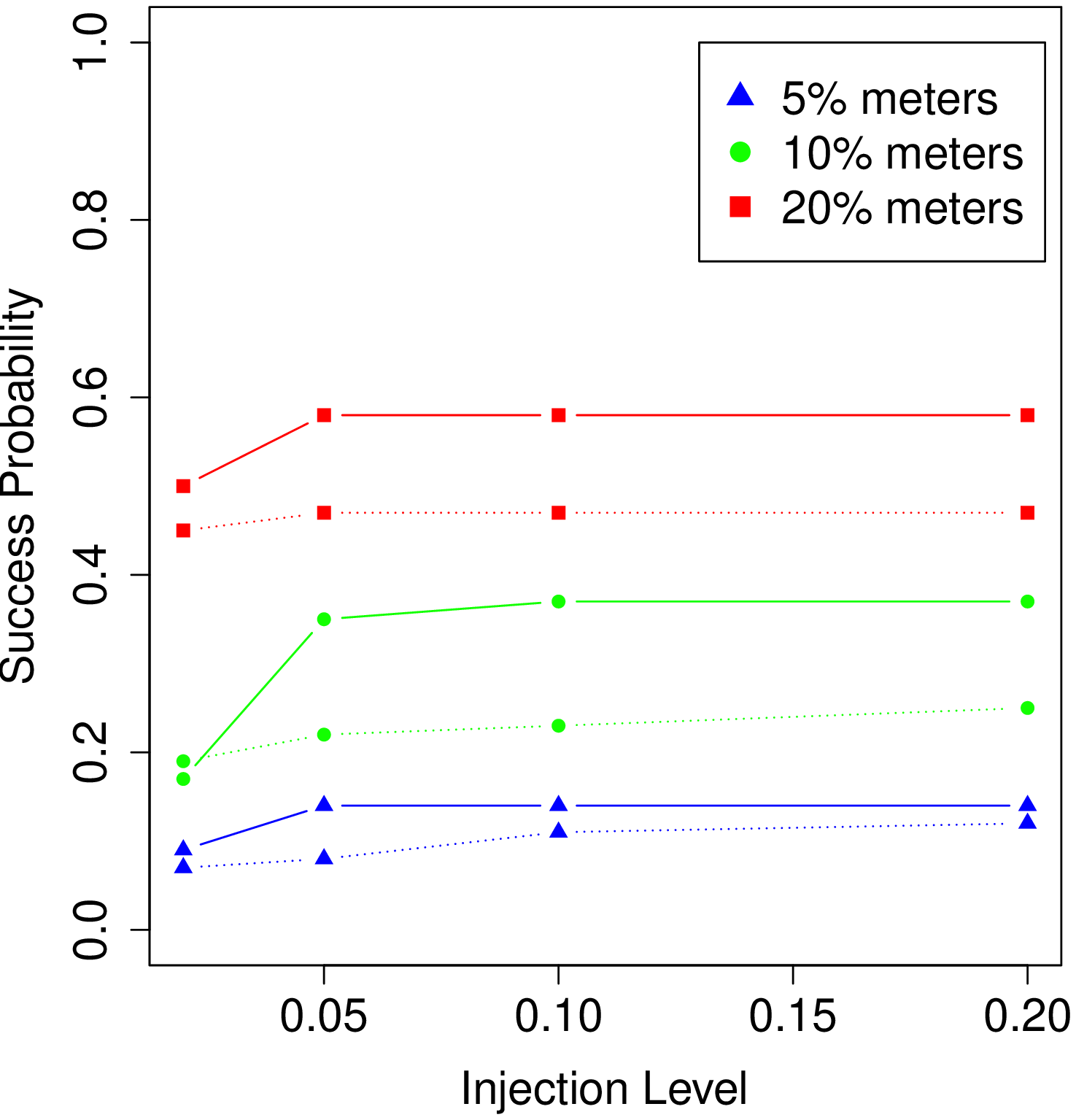}
}
\quad
\subfigure[14-bus Success Prob]{
\includegraphics[width=0.28\linewidth,height=4cm]{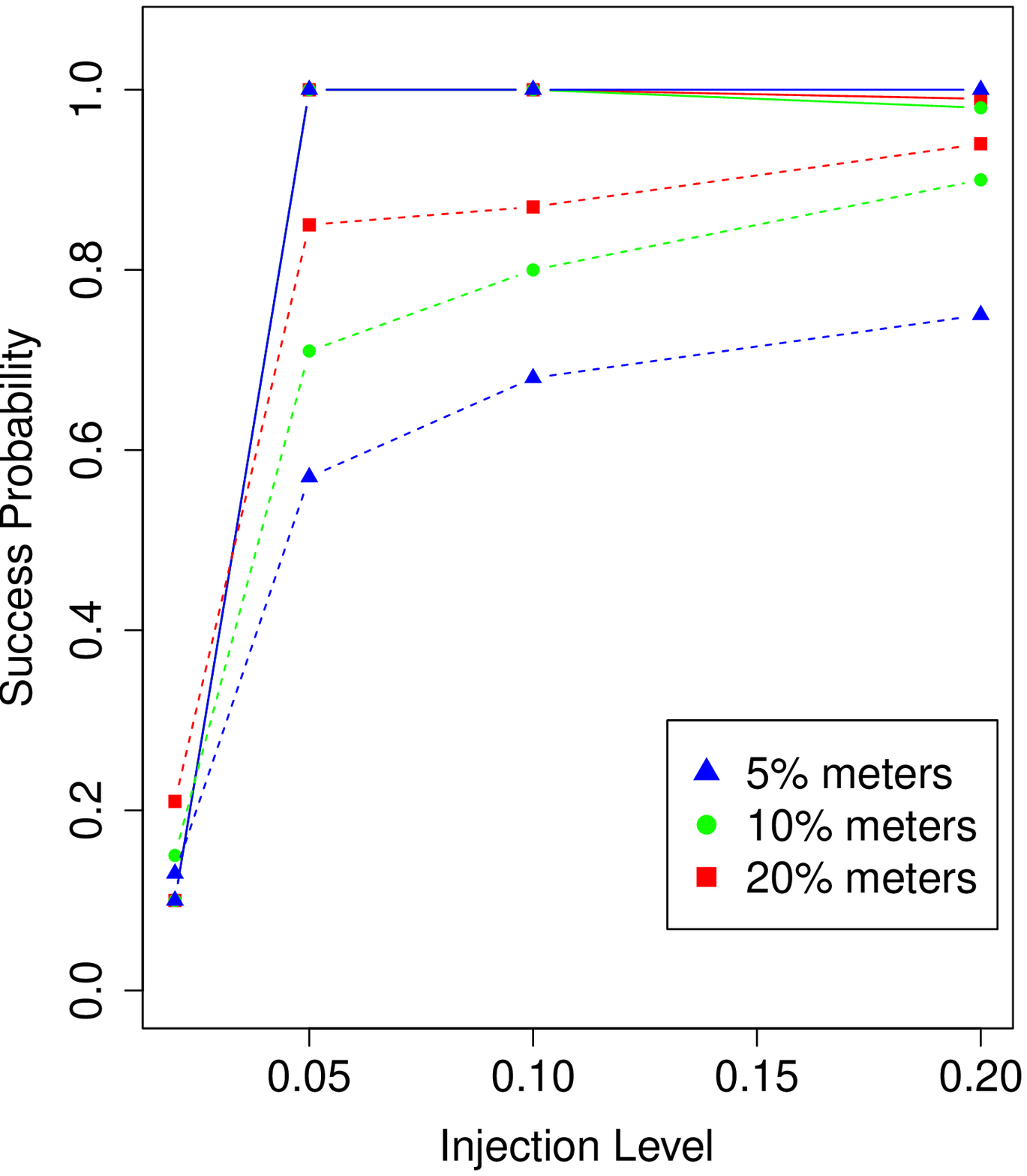}
}
\quad
\subfigure[30-bus Success Prob.]{
\includegraphics[width=0.28\linewidth,height=4cm]{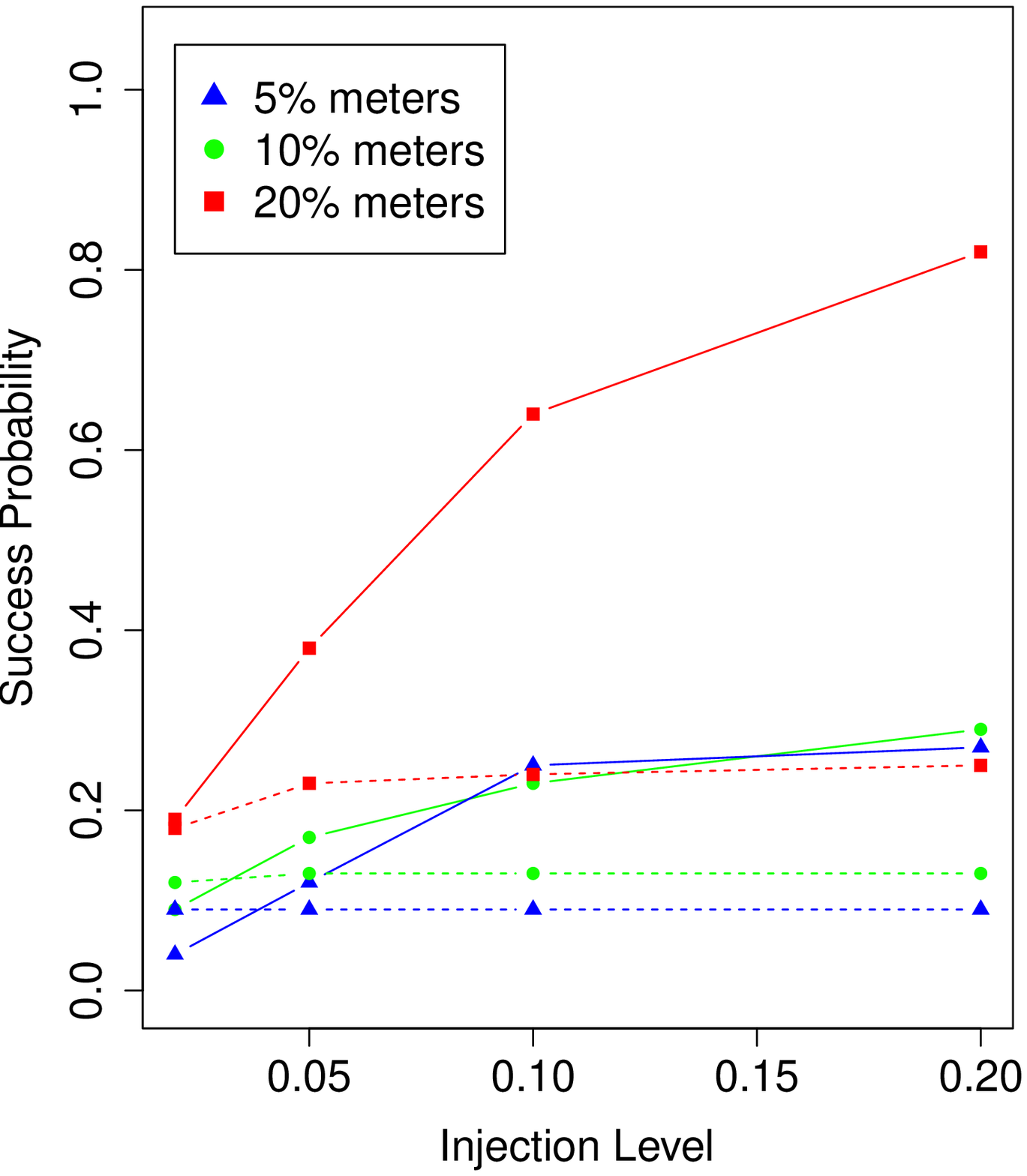}
}
\caption{Relative Error (first row) and Success Prob. (second row) of Specific $k$ Meter Attack}
\label{fig4}
\end{figure}

In general, DE algorithm outperforms the gradient-based algorithm in effectiveness (Figure~\ref{fig4}). This is not surprising, as DE brings in more diversity in every generation while SLSQP only explores neighbors in each iteration. 

Table~\ref{tab6} shows the execution time of DE attack with $1\times10^4$ NFEs and SLSQP attack with 100 iterations. Both attacks can be finished quickly within 3 seconds, which is feasible for an attacker to mount on smart grid. The comparison of running time between them can be misleading, since the execution time highly relies on NFEs for DE and max iteration numbers for SLSQP. In addition, the execution time can be further shortened by implementing a early-stop criteria or parallel processing for DE, or adjusting the max iteration numbers for SLSQP. Therefore, taking no account for running time, our experiments exhibit clear pattern that DE attack is more effective than SLSQP attack.
\vspace{-0.3cm}
\begin{table}[hbpt] 
\caption{Execution Time (in second) Comparison of Specific $k$ Attacks}
\begin{center}
\renewcommand{\arraystretch}{1.2}
\begin{tabular}{*3c }
\hline
\textbf{Test System}& \textbf{DE (s)} & \textbf{SLSQP (s)}\\
\hline
9-bus & 0.12-0.4 & 0.036-0.6 \\
14-bus & 0.06-0.6 & 0.14-1.0\\
30-bus & 0.3-3.0 & 0.26-2.2\\
\hline
\end{tabular}
\label{tab6}
\end{center}
\end{table}
\section{Conclusions}
\vspace{-0.4cm}
In this paper, we perform the first study of adversarial FDI attacks against ANN-based AC SE. We first create target models that are sufficiently strong. Then we formulate the adversarial FDI attack into an optimization problem. We extensively evaluate the proposed attacks under two attack scenarios on three test systems, with adapted DE and SLSQP aiming to find attack vectors. In the any $k$ meter attack, our results show that the DE attack is successful with high probability even with a small number of compromised meters and low false injection level. DE outperforms SLSQP in the specific $k$ meter attack. 
\vspace{-0.2cm}
\section*{Acknowledgement}
\vspace{-0.2cm}
The work of T. Shu is supported in part by NSF under grants CNS-1837034, CNS-1745254, CNS-1659965, and CNS-1460897. Any opinions, findings, conclusions, or recommendations expressed in this paper are those of the author(s) and do not necessarily reflect the views of NSF.
\bibliographystyle{splncs04}
\bibliography{main.bib}
\end{document}